\documentclass[sigconf]{acmart}
\usepackage{graphicx}
\usepackage{amssymb} % define this before the line numbering.
\usepackage{color}
\usepackage{algorithm}  
\usepackage{algorithmic}  
\usepackage{multirow}
\usepackage{bbding}

\usepackage{hyperref}
\hypersetup{
    colorlinks=true,
    linkcolor=blue,
    filecolor=blue,      
    urlcolor=blue,
    citecolor=cyan,
}

\AtBeginDocument{%
  \providecommand\BibTeX{{%
    \normalfont B\kern-0.5em{\scshape i\kern-0.25em b}\kern-0.8em\TeX}}}

\acmYear{2020}
\setcopyright{acmlicensed}\acmConference[MM '20]{Proceedings of the 28th ACM International Conference on Multimedia}{October 12--16, 2020}{Seattle, WA, USA}
\acmBooktitle{Proceedings of the 28th ACM International Conference on Multimedia (MM '20), October 12--16, 2020, Seattle, WA, USA} \acmPrice{15.00}
\acmDOI{10.1145/3394171.3413620} \acmISBN{978-1-4503-7988-5/20/10}

\acmConference[MM '20]{Proceedings of the 28th ACM International Conference on Multimedia}{October 12--16, 2020}{Seattle, WA, USA}
\acmBooktitle{Proceedings of the 28th ACM International Conference on Multimedia (MM '20), October 12--16, 2020, Seattle, WA, USA}
\acmPrice{15.00}
\acmISBN{978-1-4503-7988-5/20/10}

\begin{document}
\fancyhead{}

\title{DFEW: A Large-Scale Database for Recognizing Dynamic Facial Expressions in the Wild}

\author{Xingxun Jiang}
\authornote{Both authors contributed equally to this research.}
\email{jiangxingxun@seu.edu.cn}
\orcid{0000-0002-2139-8623}
\author{Yuan Zong}
\authornotemark[1]
\email{xhzongyuan@seu.edu.cn}
\affiliation{%
  \institution{School of Biological Science and Medical Engineering, Southeast University}
%  \streetaddress{P.O. Box 1212}
  \city{Nanjing}
%  \state{}
\country{China}
}

\author{Wenming Zheng}
\authornote{Corresponding author}
\affiliation{%
  \institution{Key Laboratory of Child Development and Learning Science, Southeast University}
%  \streetaddress{1 Th{\o}rv{\"a}ld Circle}
  \city{Nanjing}
%  \state{}
\country{China}}
\email{wenming\_zheng@seu.edu.cn}

\author{Chuangao Tang}
\affiliation{%
 \institution{School of Biological Science and Medical Engineering, Southeast University}
  \city{Nanjing}
%  \state{}
\country{China}}
\email{tcg2016@seu.edu.cn}

\author{Wanchuang Xia}
\affiliation{%
 \institution{School of Cyber Science and Engineering, Southeast University}
  \city{Nanjing}
%  \state{}
\country{China}}
\email{xiawanchuag@seu.edu.cn}

\author{Cheng Lu}
\affiliation{%
 \institution{School of Information Science and Engineering, Southeast University}
  \city{Nanjing}
%  \state{}
\country{China}}
\email{cheng.lu@seu.edu.cn}

\author{Jiateng Liu}
\affiliation{%
 \institution{School of Biological Science and Medical Engineering, Southeast University}
  \city{Nanjing}
%  \state{}
\country{China}}
\email{Jiateng_Liu@seu.edu.cn}

\renewcommand{\shortauthors}{Xingxun Jiang, Yuan Zong, and Wenming Zheng et al.}

\begin{abstract}
Recently, facial expression recognition (FER) in the wild has gained a lot of researchers' attention because it is a valuable topic to enable the FER techniques to move from the laboratory to the real applications. In this paper, we focus on this challenging but interesting topic and make contributions from three aspects. First, we present a new large-scale 'in-the-wild' dynamic facial expression database, DFEW (Dynamic Facial Expression in the Wild), consisting of over 16,000 video clips from thousands of movies. These video clips contain various challenging interferences in practical scenarios such as extreme illumination, occlusions, and capricious pose changes. Second, we propose a novel method called Expression-Clustered Spatiotemporal Feature Learning (EC-STFL) framework to deal with dynamic FER in the wild. Third, we conduct extensive benchmark experiments on DFEW using a lot of spatiotemporal deep feature learning methods as well as our proposed EC-STFL. Experimental results show that DFEW is a well-designed and challenging database, and the proposed EC-STFL can promisingly improve the performance of existing spatiotemporal deep neural networks in coping with the problem of dynamic FER in the wild. Our DFEW database is publicly available and can be freely downloaded from \url{https://dfew-dataset.github.io/}.
\end{abstract}

\begin{CCSXML}
<ccs2012>
   <concept>
       <concept_id>10003120.10003145.10011770</concept_id>
       <concept_desc>Human-centered computing~Visualization design and evaluation methods</concept_desc>
       <concept_significance>500</concept_significance>
       </concept>
   <concept>
       <concept_id>10010147.10010178.10010224</concept_id>
       <concept_desc>Computing methodologies~Computer vision</concept_desc>
       <concept_significance>500</concept_significance>
       </concept>
 </ccs2012>
\end{CCSXML}

\ccsdesc[500]{Human-centered computing~Visualization design and evaluation methods}
\ccsdesc[500]{Computing methodologies~Computer vision}

\keywords{Dynamic facial expression; Facial expression database; in-the-wild facial expression recognition; deep learning}


\maketitle

\begin{table*}[t]
\begin{center}
\caption{Summary of existing databases of dynamic facial expression in the wild.}
\label{tab:tab1}
\begin{tabular}{cccccc}
\toprule
Database & \#Sample & Source & Expression Distribution & \#Annotation Times & Available?\\
\midrule
Aff-Wild~\cite{kollias2019deep} & 298 & Web & Valence-arousal & 8 &\href{https://ibug.doc.ic.ac.uk/resources/first-affect-wild-challenge/}{Yes}\\
AFEW 7.0~\cite{dhall2019emotiw} & 1,809 & 54 Movies & 7 basic expressions & 2 & \href{https://sites.google.com/site/emotiwchallenge/}{Yes}\\
AFEW-VA~\cite{kossaifi2017afew} & 600 & AFEW database & Valence-arousal&2 &\href{https://www.baidu.com}{Yes}\\
CAER~\cite{Lee2019Context} & 13,201 & 79 TVshows & 7 basic expressions & 3 & \href{https://caer-database.github.io/}{Yes}\\
DFEW & 16,372 & 1500 movies & 7 basic expressions &10 &\href{https://dfew-dataset.github.io/}{Yes}\\
\bottomrule
\end{tabular}
\end{center}
\end{table*}

\section{Introduction}
Facial expression is one of the most naturally pre-eminent ways for human beings to communicate their emotions in daily life~\cite{darwin1998expression}. Imagine that if computers were able to understand emotions from facial expressions as human beings, our human-computer interaction (HCI) systems would be more friendly and natural. Due to this reason, facial expression recognition (FER) has become a hot research topic among HCI and multimedia analysis communities. Over the past decades, researchers have proposed a lot of well-performing methods for recognizing facial expressions, and these methods achieved promising performance in the lab-controlled environments~\cite{zheng2006facial,zhao2007dynamic,zheng2010emotion,liu2014learning,pan2019occluded,fan2020facial}. However, FER techniques are still far from the practical applications. One of the main reasons is that the facial expressions in the lab-controlled scenarios are different from the real-world ones. The unconstrained real-world facial expression often suffers from occlusions, illumination variation, pose changes, and many other unpredictable and challenging interferences, making the performance of most existing FER techniques drop sharply. For this reason, many researchers have recently shift-
\noindent ed their focus to a challenging but meaningful FER topic, i.e., \textbf{FER in the wild}, where 'in the wild' refers to the challenging conditions in unconstrained real-world environments. 

Similar to conventional FER, FER in the wild can be divided into two types of task according to the form of samples. One is static FER in the wild, whose aim is to predict the expression category from unconstrained facial images. The other is dynamic FER in the wild, in which the data describing the expression information, is the video clip or image sequence. Inspired by the success of deep learning in many vision tasks, some researchers have begun to construct large-scale facial expressions in the wild databases by resorting to the Internet that contains abundant facial expression resources. For example, Benitezquiroz et al.~\cite{benitezquiroz2016emotionet} collected facial images from the Internet and then created a large-scale static facial expression in the wild database called EmotioNet. EmotioNet includes 1,000,000 facial expression images, in which 25,000 images were manually labeled with 11 facial Action Units (AUs). Subsequently, Mollahosseini et al.~\cite{mollahosseini2019affectnet} constructed a much larger-volume database, i.e., AffectNet, consisting of 450,000 well-labeled facial image samples queried from the Internet. Recently, Li et al.~\cite{li2017reliable,li2018reliable} presented a novel static facial expression database, RAF-DB, containing nearly 30,000 web-queried facial images. Compared with EmotioNet and AffectNet, the major advantage of RAF-DB is the annotation. RAF-DB collectors hired 315 individuals as the annotators, and each sample in RAF-DB is labeled about 40 times to ensure its labeling reliability.

Unfortunately, in contrast to the static facial expressions in the wild, only a few unconstrained dynamic facial expression databases have been released until now. In the work of~\cite{dhall2012collecting}, Dhall et al. built a dynamic facial expression in the wild database, i.e., acted facial expressions in the wild (AFEW), which has been updated to the 7th version (AFEW 7.0)~\cite{dhall2019emotiw} and consists of 1,809 video clips collected from 54 movies. Recently, Lee et al.~\cite{Lee2019Context} built a large-scale benchmark for dynamic FER in the wild, called CAER, by collecting 13,201 video clips from 79 TV shows. Each clip was individually labeled by three annotators. To the best of our knowledge, CAER is the first large-scale database of dynamic facial expression in the wild. However, due to the lack of large-scale databases, the progress of deep learning methods for \textbf{dynamic FER in the wild} is seriously hindered. For example, in EmotiW2019, the annual emotion recognition challenge held at ACM ICMI based on the AFEW database, Li et al.~\cite{li2019bi} proposed a weighted fusion method integrating multiple prediction scores learned by different spatiotemporal feature learning networks, and won the champion. Nevertheless, the accuracy of the test set they achieved is only 62.78\% (7 expression classification task), which is still at a low level and does not meet the requirement of practical applications.

In order to remove the barrier of data volume to the research of dynamic FER in the wild, in this paper, we first present a new large-scale and well-annotated unconstrained dynamic facial expression database, DFEW (Dynamic Facial Expression in the Wild). DFEW can be served as a benchmark for researchers to develop and evaluate their methods for dealing with dynamic FER in the wild. To see the characteristics of DFEW, we summarize existing databases of dynamic facial expressions in the wild in Table~\ref{tab:tab1}. From Table~\ref{tab:tab1}, it can clearly be seen that our DFEW has three major advantages over existing databases including Aff-Wild~\cite{kollias2019deep}, AFEW 7.0~\cite{dhall2019emotiw}, AFEW-VA~\cite{kossaifi2017afew}, and CAER~\cite{Lee2019Context}. First, DFEW database has currently largest number of dynamic facial expression samples reaching over 16,000 video clips. Second, the forms of scene and sample in DFEW are many and varied because its video clips are collected from over 1,500 movies all over the world covering various challenging interferences, e.g., extreme illuminations, self-occlusions, and capricious pose changes. Last but not least, each sample in DFEW has been individually labeled ten times by the annotators under professional guidance. 

In addition to DFEW, we also propose a novel method called Expression-Clustered Spatiotemporal Feature Learning (EC-STFL) framework to deal with dynamic FER in the wild. EC-STFL framework can enforce the spatiotemporal deep neural networks, e.g., C3D~\cite{tran2015learning} and P3D~\cite{qiu2017learning}, to better learn discriminative features describing dynamic facial expressions in the wild. Finally, we establish a benchmark evaluation protocol for DFEW and conduct extensive experiments using many spatiotemporal deep learning methods as well as our proposed EC-STFL. Experimental results show that the proposed EC-STFL framework can promisingly improve the performance of existing spatiotemporal neural networks in coping with FERW problem.

\begin{figure}[h]
\centering 
\includegraphics[width=8.2cm]{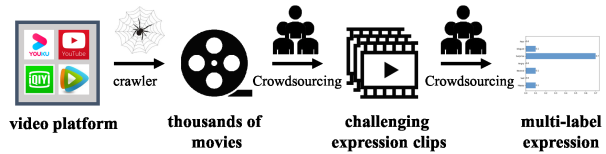}   %\columnwidth
\caption{Overview of the construction and the annotation of DFEW.}
\label{fig:pipeline}
\end{figure}

\begin{figure*}[t]
\centering
\includegraphics[width=\textwidth]{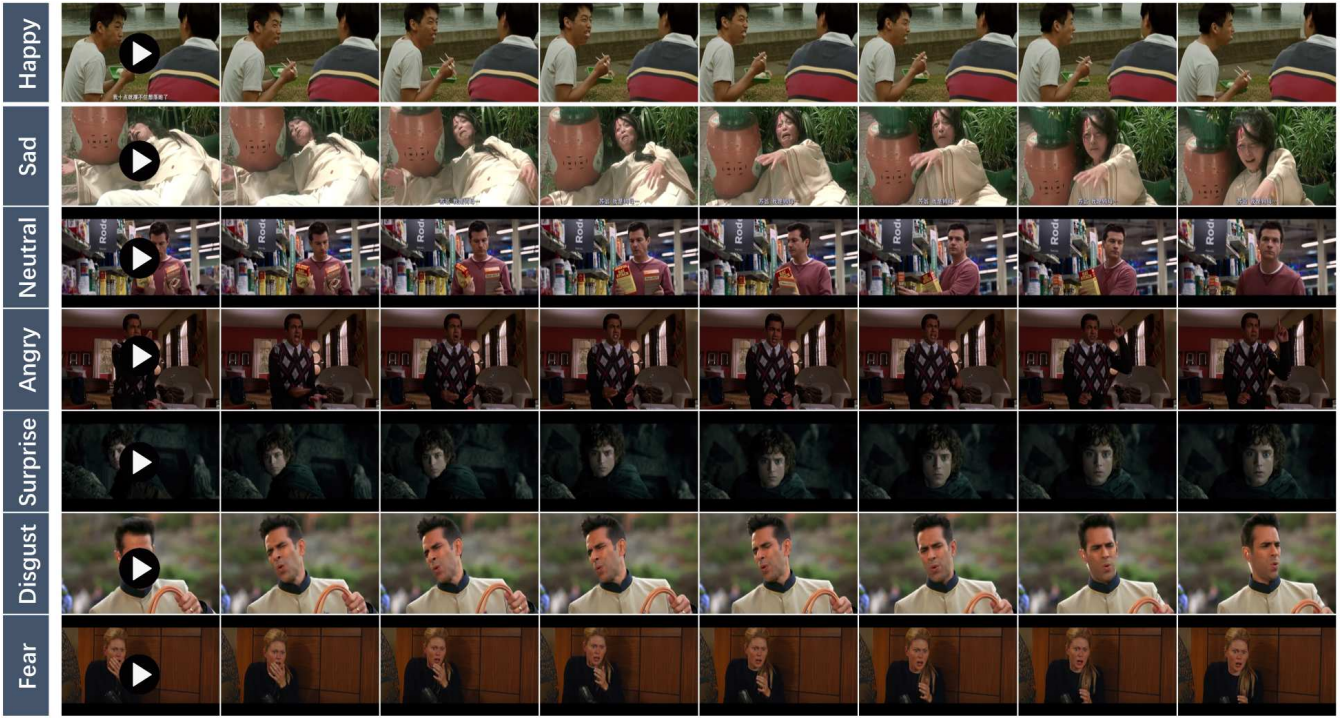}
\caption{Examples of seven basic emotions from single-labeled DFEW.}
\label{fig:databaseDemo}
\end{figure*}

\section{DFEW database}
\subsection{Data Collection}
It is believed that movies originate from and mimic our real life, hence actresses and actors in movies may have all kinds of unconstrained facial expressions originally existing in the practical scenarios. Thus it offers us abundant samples of dynamic facial expressions. By extracting the video clips containing different facial expressions from movies, we are able to build a large-scale database of dyanmic facial expressions in the wild. Following this method, several dynamic facial expression databases, e.g., Aff-Wild~\cite{kollias2019deep}, AFEW~\cite{dhall2012collecting,dhall2019emotiw}, and CAER~\cite{Lee2019Context} have been successively built and released over the past few years, which indeed advances the research of dynamic FER in the wild. In this paper, we also take full advantage of movies to collect unconstrained dynamic facial expression samples to build our DFEW database.

The pipeline of building the DFEW database is shown in Fig.~\ref{fig:pipeline}. As Fig.~\ref{fig:pipeline} shows, we first make use of crawler to collect over 1,500 high-definition movies close to our real life and covering various themes, e.g., comedy, tragedy, war, and love, from the Internet to serve as the sample source of facial expressions in the wild. Then, we hired dozens of students to use video editing software to manually extract video clips containing one of seven basic expressions from their assigned movies. Note that we made several rules to help these student extractors ensure the diversity of their extracted facial expression samples. For example, the students are only allowed to extract at most 20 video clips from each movie. Meanwhile, an additional reward would be given to one student if he or she submitted the samples of relatively rare facial expressions, e.g., disgust and fear. Through the above method, we ultimately collected 16,372 unconstrained facial expression video clips.

\subsection{Data Annotation}
High-quality data annotation is another challenge for the database. First of all, annotating such a large database is time-consuming and needs efficient personnel management. Second, though psychologists P. Ekman believes that the seven basic emotions are universal and independent of the cultural mismatch~\cite{ekman1971constants}, culture mismatch indeed exists and worth considering because the labeling bias can be removed as far as possible. To efficiently manage annotators and understand the protagonist's emotion in clips better, we entrust the labeling work to the professional crowdsourcing company, JD crowdsourcing~\footnote{http://weigong.jd.com/}, where we hired twelve expert annotators. They are asked to identify each clip's closest emotion in seven typical discrete emotions, i.e., anger, disgust, fear, happy, sad, surprise, and neutral. Before formal annotation, these twelve annotators are professionally trained with the emotional knowledges. Then each clip is annotated by ten independent annotators. After annotation, we obtained the seven-dimensional emotion vectors or emotion distribution annotating information of 16,372 clips.

We suppose the seven-dimensional emotion ground truth of $j$-th video clip denoted by $L_j = \{ l_1,..., l_k, …, l_7 \}$, where $l_k$ represents the annotation times of $k$-th emotion labeled by annotators, $k \in \{1,2,3,4,5,6,7\}$ refer to happy, sad, neutral, angry, surprise, disgust and fear, respectively. 

 \begin{table}[h]
 \begin{center}
 \caption{The basic information of single-labeled DFEW.}
 \label{tab:basicInfoDFEWTypical}
 \begin{tabular}{ccccccc}
 \toprule
 \multirow{2}{*}{Emotions}&\multicolumn{4}{c}{Clips} & \multirow{2}{*}{Percent} \\
     \cline{2-5}
     & 0-2s  & 2-5s & 5s+  & Total &\\
 \midrule
 Happy        &  852  & 1252 &  384 & 2488  & 20.63\\
 Sad          &  440  &  915 &  653 & 2008  & 16.65\\
 Neutral      &  832  & 1335 &  542 & 2709  & 22.46\\
 Angry        &  762  & 1091 &  376 & 2229  & 18.48\\
 Surprise     &  691  &  648 &  159 & 1498  & 12.42\\
 Disgust      &   71  &   58 &   17 &  146  &  1.22\\
 Fear         &  408  &  435 &  138 &  981  &  8.14\\
 \midrule
 Total        & 4056  & 5734 & 2269 & 12059 & 100.00 \\
 \bottomrule
 \end{tabular}
 \end{center}
 \end{table}

However, not all clips can be further clearly assigned to a specific single-labeled emotion category from multi-dimensional emotion distribution. Therefore, for accurate labeling, we pick out the emotion $k$ as the single label with respect to $l_k > r$, where $r$ is the threshold value of annotation times. In this work, we set the threshold value $r=6$, hence select 12059 clips of DFEW to be the single-labeled. We provide basic information of single-labeled DFEW in Table~\ref{tab:basicInfoDFEWTypical}, and demo samples of single-labeled DFEW in Fig.~\ref{fig:databaseDemo}. Note that, to promote emotion research, we will release both single-labeled annotation and seven-dimensional emotion distribution annotation.

\subsection{Agreement Test}
In this section, we discuss the quality of emotion annotation based on Fleiss's Kappa test~\cite{fleiss1971measuring}. Fleiss's Kapaa test calculates the degree of agreement in classification over that which would be expected by chance. We believe that its result is an excellent index to give annotation's reliability or quality. In the task of annotating clips, ten independent individuals annotate each clip  with $k \in \{ 1,2,3,4,5,6,7 \}$, i.e., one of the seven typical discrete emotions. Here, we let $n_{ij}$ represent the number of annotators who assigned the $i$-th clip to the $j$-th emotion. So we can calculate $p_j$, the proportion of all assignments which were to the $j$-th emotion,

\begin{align}
\left \{
    \begin{array}{l}
    p_j=\frac{1}{N \times n} \sum_{i=1}^N \limits n_{ij}\\
    \sum_{j=1}^K \limits p_j=1
    \end{array}
\right.
\end{align}
\noindent where $n=10$ is the annotation time of each clips, $K=7$ is the number of emotion category, and $N$ is the number of clips. And we can calculate $P_i$, the extent to which annotators agree for the $i$-th clip, i.e., compute how many annotator-annotator pairs are in agreement, relative to the number of all possible annotator--annotator pairs:

\begin{align}
P_i=\frac{1}{n \times (n-1)} \left[ (\sum_{j=1}^K \limits n_{ij}^2) - n \right]
\end{align}

And compute $\bar{P}$, the mean of $P_i$, and $\bar{P_e}$ which go into the formula for coefficient $\kappa$:

\begin{align}
\bar{P} = \frac{1}{N} \sum_{i=1}^{N} \limits P_i
\end{align}

\begin{align}
\bar{P_e} = \sum_{j=1}^{k} \limits p_j^2
\end{align}

Then we can calculate $\kappa$ by 

\begin{align}
\kappa = \frac{\bar{P}-\bar{P_e}}{1-\bar{P_e}}
\end{align}

We perform Fleiss's Kapaa test both in the whole DFEW database and the single-labeled part, and we obtain $\kappa=0.70$ for the whole DFEW database and $\kappa=0.63$ for the single-labeled part. Based on Table~\ref{tab:kapaaInterpretation}, we believe that all annotators achieve a substantial agreement. That is to say, our annotation is of high quality.

\begin{table}[h]
\begin{center}
\caption{Interpretation of $\kappa$ for Fleiss' Kapaa Test.}
\label{tab:kapaaInterpretation}
\begin{tabular}{cc}
\toprule
$\kappa$&Interpretation\\
\midrule
\textless0&Poor agreement\\
0.01-0.20&Slight agreement\\
0.21-0.40&Fair agreement\\
0.41-0.60&Moderate agreement\\
0.61-0.80&Substantial agreement\\
0.81-1.00&Almost perfect agreement\\
\bottomrule
\end{tabular}
\end{center}
\end{table}

\section{Expression-Clustered Spatiotemporal Feature Learning}
The challenge of dynamic FERW is how to learn robust and discriminative features to describe facial expression video clips, the facial expression representation of video clips, which are contaminated by the abnormal conditions, such as variations of illumination, posture, occlusion and scale. Spatiotemporal features obtained by the various spatiotemporal neural networks are adept in characterizing the dynamic face motion in video samples from the spatial stream and temporal stream. Because of the strong fitting ability of neural networks, the hierarchical spatiotemporal features perform better than the traditional methods in the anti-noise problem. Unfortunately, the margin of different emotion features distributed in the feature space is still blurring due to those abnormal or challenging conditions. To simultaneously cope with FERW and the make feature margins clear, we propose an Expression-Clustered Spatiotemporal Feature Learning (EC-STFL) framework, which can be embedded in the popular spatiotemporal network flexibly. Drawing on the idea of LDA, the EC-STFL enhances intra-class correlation and reduces inter-class correlation by designing special similarity matrices, and is formulated as follows,

\begin{align}
\min\limits_{W}  \sum_{i,j} \frac{P_{ij} \phi(x_i, x_j)}{Q_{ij} \phi(x_i, x_j)}
\end{align}

\noindent where $W$ is the network's weight, matrix $P$ and matrix $Q$ are both similarity matrices, $\phi(x_i, x_j) = \left\| x_i - x_j \right\|$ is the spatiotemporal feature distance of sample $x_i$ and sample $x_j$, where $x \in \mathbb{R}^d$ is extracted from the final hidden fully connected layers, i.e., just before the softmax layer that produces the class prediction. And the matrix $P$ and matrix $Q$ are defined as follows:

\begin{align}
P_{ij} =
  \begin{cases}
    0, & \text{if $x_i$ and $x_j$ has the same label}\\
    1, & \text{otherwise}
  \end{cases}
\end{align}

\begin{align}
Q_{ij} =
  \begin{cases}
    0, & \text{if $x_i$ and $x_j$ has the different label}\\
    1, & \text{otherwise}
  \end{cases}
\end{align}

Obviously, the EC-STFL minimizes the feature distance between the same emotions and maximize the feature distance between different emotions to clarify the emotion margin in spatiotemporal feature space. To implement it more effectively and efficiently, we calculate EC-STFL loss in the mini-batch because of limited memory. Besides, we note that sample unbalance widely exists in the FER task~\cite{li2019bi,liu2018multi,hu2017learning,fan2016video}, which leading the classifiers prefer the emotions with more samples and ignoring the emotions with fewer samples. The FER task in our DFEW database also faces this trouble. Considering that, we develop the EC-STFL loss by adding dynamic weights to balance different emotions’ loss in the update progress of batch loss, and extend EC-STFL loss as follows,

\begin{align}
  L_{EC-STFL} = \frac{\sum\limits_{1 \leq i, j \leq n, x_j\in\mathcal{N}\{x_i\}} \frac{\left\| x_i - x_j \right\|}{N_{x_i}}}{\sum\limits_{1 \leq i, j \leq n, x_j\notin\mathcal{N}\{x_i\}} \frac{\left\| x_i - x_j \right\|}{N_{x_j}}}
\end{align}

\noindent where $ \mathcal{N}\{x_i\} $ is the set of the same single-labeled emotion annotation with $x_i$ in mini-batch, $N_{x_i}$ is the set size of $ \mathcal{N}\{x_i\}$, and $n$ is the mini-batch size. Creating the dynamic weights by $N_{x_i}$ and $N_{x_j}$, EC-STFL adjusts and balances the losses of different emotions in each mini-batch, hence alleviate the imbalance issue of FER task to some degree.

We adopt joint supervision for training softmax loss and our EC-STFL loss to obtain the discriminative spatiotemporal features. The total objective function expressed as $L = L_s + \lambda L_{EC-STFL}$, where $L_s$ denotes softmax loss and hyper-parameter $\lambda$ is a coefficient used to trade-off $L_s$ and $L_{EC-STFL}$. Note that, we drop the backward step when $L_{EC-STFL}$ has no meaning, i.e., mini-batch only contains samples with one kind of emotion.

\section{Experiments}
In this section, we give an experimental setup for benchmark first, including data preprocessing, experimental protocol, and evaluation metric. Then we conduct extensive spatiotemporal neural network methods for the investigations of our DFEW database, and these networks with EC-STFL loss for the verification. Finally, we make transfer experiments from some widely used action databases and our DFEW database to AFEW database, to verify DFEW can extract adequate and efficient transfer knowledge for the FERW task.

\subsection{Experimental Setup}
\textbf{Data\&Protocol.}
To better evaluate the single-labeled DFEW database with 12,059 video clips, we adopt a 5-fold cross-validation protocol for the benchmarks, which means we split all the samples into five same-size parts without overlap to conduct experiments. In each fold (fd1$~\sim$ fd5), one part of samples are used for testing and the remaining for training. Finally, all the predicted labels are used to compute the evaluation metrics by comparing the ground truth.

\textbf{Preprocessing.}
First, we use OpenCV to extract image frames from 12,059 clips, face++ API~\cite{faceplusplusapi} to acquire face region images and facial landmarks. We remove the non-face (undetected) frames and statistics the useful frame rate of clips to eliminate those less than 50\%. Totally 362 clips were not taken into consideration. Then, we use SeetaFace~\cite{liu2016viplfacenet} for face affine transformation, which normalizes faces based on acquired facial landmarks. Finally, we align temporal length of the remaining clip samples into 16 frames using the time interpolation method in ~\cite{zhou2011towards,zhou2013compact}. 

\begin{table*}[t]
\begin{center}
\caption{Comparsion of the seven basic emotion classification performance of C3D, P3D, R3D18, 3D Resnet18, I3D-RGB, VGG11+LSTM, Resnet18+LSTM on DFEW database. The metrics include UAR(unweighted average recall) and WAR(weighted average recall).}
\label{tab:baselineScratch}
\begin{tabular}{ccccccccccc}
\toprule
\multirow{2}{*}{Model}&\multicolumn{7}{c}{Emotions} &~& \multicolumn{2}{c}{Metric} \\
    \cline{2-8} \cline{10-11}
                                             &  Happy & Sad & Neutral & Angey & Surprise & Disgust & Fear & ~ & UAR & WAR\\
\midrule
C3D~\cite{tran2015learning}                  & 75.17  &39.49& 55.11   & 62.49 & 45.00    & 1.38    & 20.51& ~ & 42.74       & 53.54 \\
P3D~\cite{qiu2017learning}                   & 74.85  &43.40& 54.18   & 60.42 & \textbf{50.99 }   & 0.69    & 23.28& ~ & 43.97       & \textbf{54.47} \\
R3D18~\cite{tran2018closer}                  & \textbf{79.67}  &39.07& 57.66   & 50.39 & 48.26    & 3.45    & 21.06& ~ & 42.79       & 53.22 \\
3D Resnet18~\cite{hara2017learning}          & 73.13  &\textbf{48.26}& 50.51   & \textbf{64.75} & 50.10    & 0.00    & \textbf{26.39}& ~ & \textbf{44.73}       & 54.98 \\
I3D-RGB~\cite{carreira2019short}             & 78.61  &44.19& 56.69   & 55.87 & 45.88    & 2.07    & 20.51& ~ & 43.40       & 54.27 \\
VGG11+LSTM~\cite{simonyan2014very,hochreiter1997long,gers1999learning}& 76.89  &37.65& \textbf{58.04}   & 60.70 & 43.70    & 0.00    & 19.73& ~ & 42.39       & 53.70 \\
Resnet18+LSTM~\cite{he2016deep,hochreiter1997long,gers1999learning}& 78.00  &40.65& 53.77   & 56.83 & 45.00    & \textbf{4.14}    & 21.62& ~ & 42.86       & 53.08 \\
\bottomrule
\end{tabular}
\end{center}
\end{table*}

\textbf{Evaluation Metric.}
We choose two metrics~\cite{schuller2010cross} widely used in existing researches for evaluating the unbalanced problems, i.e., the unweighted average recall (UAR, i.e., the accuracy per class divided by the number of classes without considerations of instances per class) and weighted average recall (WAR, i.e., accuracy). They are appropriate for the FERW task. The UAR metric indicates the average accuracy of different facial expressions, and we can adequately evaluate the performance of predicting emotions with few samples using the UAR results. The WAR metric indicates the recognition accuracy of overall expressions. We hope to improve models' performance both in  UAR and WAR metrics.

\textbf{Implementation Details.}
In this paper, we employ the PyTorch framework~\cite{paszke2017automatic} to implement all models. All models are trained on 12G memory's Titan Xp with an excellent initial learning rate provided by the grid search strategy. And the learning rate reduced at a rate of 10$\times$ when loss saturated. 
First, we train models from scratch to present the benchmarks. Batch size is set to 24, which is the max operational batch size of C3D~\cite{tran2015learning} on Titan Xp. We set trade-off coefficient $\lambda$ of models with EC-STFL to 10, and trade-off coefficient of center loss to $1 \times 10^{-4}$ according to ~\cite{wen2016discriminative}. Second, we further discuss EC-STFL about the batch size and trade-off coefficient $\lambda$ based on C3D~\cite{tran2015learning} and 3D Resnet18~\cite{hara2017learning}. These experiments are conducted on two Titan Xp. Third, we make cross-database transfer experiments. We finetune some off-the-shelf models initilized by weights provided by other researchers with the best learning rate.

\subsection{Experimental Results}
\textbf{Baseline System.}
The existing spatiotemporal neural networks based on RGB frames can be mainly categorized into two groups: the 3D convolutional neural networks and CNN-RNN networks. In this paper, we conduct five 3D CNN models, i.e., C3D~\cite{tran2015learning}, I3D-RGB~\cite{carreira2019short}, R3D18~\cite{tran2018closer}, 3D Resnet18~\cite{hara2017learning}, P3D~\cite{qiu2017learning}, and two CNN-RNN models, i.e., VGG11+LSTM and Resnet18+LSTM for benchmarks. VGG11~\cite{simonyan2014very} and Resnet18~\cite{he2016deep} are slightly modified to fit the input size of 112 $\times$ 112. The classification results are shown in Table~\ref{tab:baselineScratch}.

\begin{table}[h]
\begin{center}
\caption{Expression recognition performance of different methods with and without EC-STFL on DFEW database.}
\label{tab:DFEWwithECSTFL}
\begin{tabular}{lcc}
\toprule
\multirow{2}{*}{Model}& \multicolumn{2}{c}{Metric}\\
    \cline{2-3}
                                & UAR  & WAR\\
\midrule
C3D                             & 42.74        & 53.54       \\
\textbf{C3D,EC-STFL}            &\textbf{45.10}&\textbf{55.50}\\
\midrule
P3D                             & 43.97        & 54.47        \\
\textbf{P3D,EC-STFL}          &\textbf{45.22}&\textbf{56.48}\\
\midrule
R3D18                           & 42.79        & 53.22        \\
\textbf{R3D18,EC-STFL}          &\textbf{45.05}&\textbf{56.19}\\
\midrule
3D Resnet18                     & 44.73        & 54.98        \\
\textbf{3D Resnet18,EC-STFL}    &\textbf{45.35}&\textbf{56.51}\\
\midrule
I3D-RGB                         & 43.40        & 54.27        \\
\textbf{I3D-RGB,EC-STFL}        &\textbf{45.05}&\textbf{56.19}\\
\midrule
VGG11+LSTM                      & 42.39        & 53.70        \\
\textbf{VGG11+LSTM,EC-STFL}     &\textbf{44.78}&\textbf{56.25}\\
\midrule
Resnet18+LSTM                   & 42.86        & 53.08        \\
\textbf{Resnet18+LSTM,EC-STFL}  &\textbf{43.60}&\textbf{54.72}\\
\bottomrule
\end{tabular}
\end{center}
\end{table}

It is seen from Table~\ref{tab:baselineScratch} that P3D~\cite{qiu2017learning} achieves the best WAR at 54.47\%, and 3D Resnet18~\cite{hara2017learning} achieves the best UAR at 44.73\% among all networks. It is an interesting finding that both UAR and WAR attained by 3D CNN models instead of CNN-RNN models. Among seven types of emotions, 3D CNN better predicts happy, sad, angry, surprise, and fear emotions, while CNN-RNN models better at neural and disgust emotions. One possible reason is that models learn feature existing preference. From Table~\ref{tab:baselineScratch}, we can also find that it is easier to classify the happy emotion while harder to the disgust. We can also find that happy emotion is more comfortable to be classified while the disgust is much harder to be well predicted. It may result from the relatively low variance of intra-class facial features for the happy emotion while significant variance for the disgust emotion, or fewer samples of the disgust. In fact, fewer disgust samples mean more serious imbalance problem, which is a widely existed problem leading the lousy performance. To the best of our knowledge, the recognition of disgust emotion is really a hard problem in the FERW task.

\textbf{EC-STFL.}
To acquire more discriminative features, we design the EC-STFL and incorporate it with some off-the-shelf 3D convolutional neural networks and CNN-RNN networks. The experiment results with and without EC-STFL are detailed in Table~\ref{tab:DFEWwithECSTFL}. We can find that all EC-STFL based models show better recognition performance than those without this module. Our EC-STFL can promote the UAR and WAR by an average of 1.61 percentage points and 2.08 percentage points, respectively. What is more, comparing with the other models from Table~\ref{tab:DFEWwithECSTFL}, we can find that 3D Resnet18 with EC-STFL achieves the best UAR and WAR results.

\begin{figure}[h]
\centering
\includegraphics[width=\columnwidth]{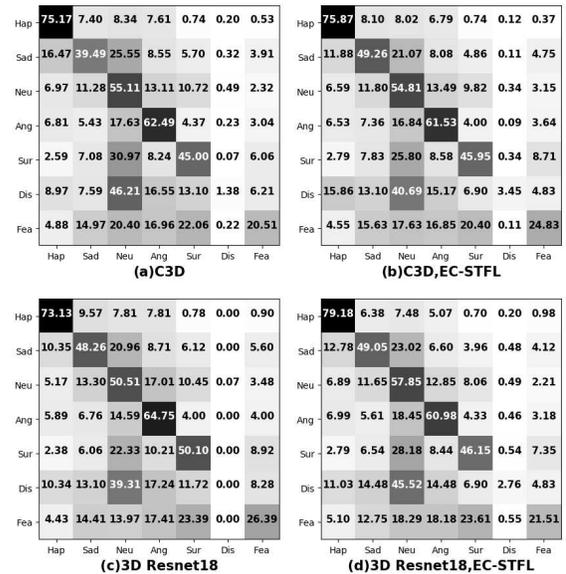}
\caption{The confusion matrices of selected methods with and without EC-STFL. (a)C3D, (b)C3D with EC-STFL, (c)3DResnet18, (d)3D Resnet18 with EC-STFL.}
\label{fig:cm2}
\end{figure}

We provide the recognition performance of different emotion detailed by confusion matrices in Fig.~\ref{fig:cm2}, to further discuss classification differences between models with and without EC-STFL. Displayed in Fig.~\ref{fig:cm2}, EC-STFL improves the recall rates of the C3D model for happy, sad, surprise, disgust, and fear emotion by 0.7\%, 9.77\%, 0.95\%, 2.07\%, and 4.32\%, respectively. EC-STFL improves the recall rate of the 3D Resnet18 model for happy, sad, neutral, disgust by 6.05\%, 0.79\%, 7.34\%, 2.76\%, respectively. Results are given in Fig.~\ref{fig:cm2} show that our EC-STFL both improve the recall rate of happy, sad, disgust for the C3D and 3D Resnet18.

\begin{figure}[h]
\centering
\includegraphics[width=\columnwidth]{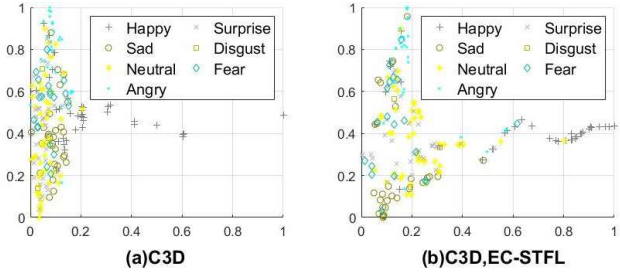}
\caption{The distribution of deeply features in (a) C3D and (b) C3D with EC-STFL, whose feature dimension is reduced by tSNE. As can be seen, EC-STFL helps the learned features more discriminative.}
\label{fig:fea}
\end{figure}

For a better understanding of the learned features by EC-STFL, we utilize a non-linear mapping method, i.e., t-SNE~\cite{maaten2008visualizing,van2014accelerating}, to visualize the learned features on a 2D plane, as shown in Fig.~\ref{fig:fea}. Compared with the models have no EC-STFL module, we observe that the features learned by EC-STFL show the more significant inter-class distance between different classes; hence the samples show a better aggregation effect. It suggests that our proposed EC-STFL has the ability to promote better feature representation.

The competitor of EC-STFL is mainly the loss inspired by the idea of clustering, e.g., the well-known “center loss”~\cite{wen2016discriminative}. In this paper, we conduct the comparsion experiments based on two spatiotemporal models, i.e., C3D and 3D Resnet18. Table~\ref{tab:ECSTFLcenterloss} contains the comparsion of center loss and EC-STFL. As is evident from the Table~\ref{tab:ECSTFLcenterloss} that EC-STFL and center loss are both improve the classification performance of models purely use cross entropy loss. Furthermore, the EC-STFL performs better than center loss, and achieves the best UAR and WAR.

\begin{table*}[h]
\begin{center}
\caption{Comparison of EC-STFL and center loss on DFEW database.}
\label{tab:ECSTFLcenterloss}
\begin{tabular}{cccccccccccl}
\toprule
\multirow{2}{*}{Model}&\multicolumn{7}{c}{Emotions} &~& \multicolumn{2}{c}{Metric} \\
    \cline{2-8} \cline{10-11}
                            &Happy&Sad&Neutral&Angry&Surprise&Disgust&Fear&~&UAR&WAR\\
\midrule
C3D                         &75.17&39.49&55.11&62.49&45.00&1.38&20.51&~&42.74&53.54\\
C3D, center loss            &75.62&44.67&54.18&63.14&42.21&2.07&22.17&~&43.44&54.17\\
\textbf{C3D,EC-STFL}        &75.87&49.26&54.81&61.53&45.95&3.45&24.83&~&\textbf{45.10}&\textbf{55.50}\\
\midrule
3D Renset18                 &73.13&48.26&50.51&64.75&50.10&0.00&26.39&~&44.73&54.98\\
3D Resnet18, center loss    &78.49&44.30&54.89&58.40&52.35&0.69&25.28&~&44.91&55.48\\
\textbf{3D Resnet18,EC-STFL}&79.18&49.05&57.85&60.98&46.15&2.76&21.51&~&\textbf{45.35}&\textbf{56.51}\\
\bottomrule
\end{tabular}
\end{center}
\end{table*}

\begin{figure}[h]
\centering
\includegraphics[width=\columnwidth]{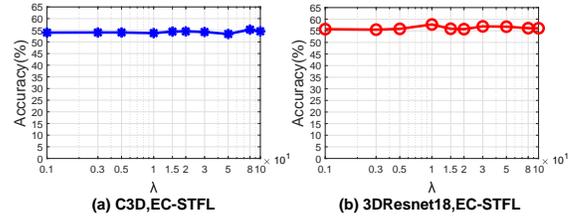}
\caption{The sensitive experiments results of trade-off parameter for the proposed EC-STFL framework. (a) C3D with EC-STFL, (b) 3D Resnet18 with EC-STFL. The scale of trade-off parameter is $\lambda \in \{ 1,3,5,10,15,20,30,50,80,100\}$.}
\label{fig:hpECSTFL}
\end{figure}

\begin{figure}[h]
\centering
\includegraphics[width=\columnwidth]{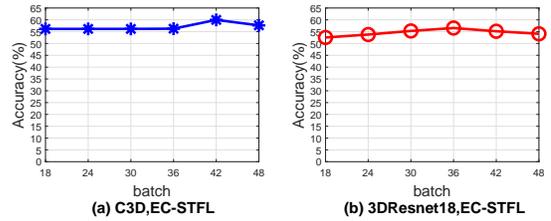}
\caption{The sensitive experiments results of batch size for the proposed EC-STFL framework. (a) C3D with EC-STFL, (b) 3D Resnet18 with EC-STFL. The scale of batch size is $m \in \{18, 24, 30, 36, 42, 48\}$.}
\label{fig:ECSTFLbatch}
\end{figure}

\textbf{Hyper-parameters Discussion.} 
The trade-off hyperparameter $\lambda$ and batch size $m$ affect the performance of EC-STFL, which are both essential to EC-STFL. So we conduct experiments to evaluate models' sensitiveness based on C3D and 3D Resnet18 in the fd1 data split. In the first experiment, we fix batch size $m=24$ and vary $\lambda\in\{1,3,5,10,15,20,30,50,80,100\}$. It is apparent that properly choosing the value of $\lambda$ can improve the verification accuracy of the learned features. In the second experiment, we fix $\lambda=10$ and vary batch size $m\in\{18,24,30,36,42,48\}$. The WAR or accuracy results are visible in Fig.~\ref{fig:hpECSTFL} and Fig.~\ref{fig:ECSTFLbatch}, respectively. Likewise, the verification performance of EC-STFL based models remain largely stable across a wide range of batch sizes.

\subsection{Transfer Learning}
We hypothesize that the DFEW database would contribute to clip-based emotion classification models' transfer learning performance on real-life applications. To verify this hypothesis, we conduct extensive transfer learning experiments from widely used action data-
\noindent bases and our DFEW database to the AFEW~\cite{dhall2012collecting} database. The action databases include UCF101~\cite{soomro2012ucf101}, Sports 1M~\cite{karpathy2014large}, Kinect 700~\cite{carreira2019short}, and Moments In Time~\cite{monfort2019moments}. We select two spatiotemporal neural networks and their EC-STFL version, i.e., C3D, 3D Resnet18, and C3D with EC-STFL, 3D Resnet18 with EC-STFL.

\begin{table*}
\begin{center}
\caption{The transfer learning performance on AFEW7.0.}
\label{tab:Transfer}
\begin{tabular}{lcccc}
\toprule
\multirow{2}{*}{Pretrained}&\multicolumn{4}{c}{Finetuned models} \\
    \cline{2-5}
 &C3D  & C3D, EC-STFL & 3D Resnet18 & 3D Resnet18, EC-STFL\\
\midrule
Sports 1M  &41.78&44.91&-&-\\
UCF101     &41.25&42.34&-&-\\
Kinect700  &-&-&49.35&49.61\\
Kinect700+Moments In Time &-&-&49.35&49.35\\
DFEW, fd2  &\textbf{44.91}&\textbf{45.56}&\textbf{53.00}&\textbf{53.26}\\
DFEW, fd5  &\textbf{49.87}&\textbf{49.87}&\textbf{49.61}&\textbf{49.66}\\
\bottomrule
\end{tabular}
\end{center}
\end{table*}

We initialize models with the corresponding pre-trained weights trained from action databases provided by other researchers and our DFEW database respectively, for example, C3D and C3D with EC-STFL use pre-trained weights of C3D model. Then finetune all the layers of network on the AFEW database at a best learning rate searched by grid strategy. Note that, we choose models' pre-trained weights on our DFEW database based on the second data split and the fifth data split, denoted by fd2 and fd5 for short, respectively. We use WAR metric as the evaluation and show the transfer results in Table~\ref{tab:Transfer}. We found that initial weights provided by the DFEW database show a better transfer learning performance than the action databases. We further compare our transfer results with those state-of-the-arts methods. As results illustrated in Table~\ref{tab:compareSOTA}, transferred 3D Resnet18 improve the state-of-the-art method on WAR about 2 percent. In this way, we can conclude that our DFEW database is useful for developing excellent emotion prediction models in real-life applications.

\begin{table}
\begin{center}
\caption{Comparison of 3D Resnet18 model’s transfer results with other state-of-the-art methods on AFEW7.0.}
\label{tab:compareSOTA}
\begin{tabular}{lcccccc}
\toprule
Model& WAR \\
\midrule
Lu et al.~\cite{lu2018multiple}&45.31\\ %18_3
Fan et al.~\cite{fan2016video}&45.43\\%16_1
Hu et al.~\cite{hu2017learning}&46.48\\ %17_1
Fan et al.~\cite{fan2018video}&48.04\\ %18_2
Liu et al.~\cite{liu2018multi}&51.44\\ %18_1
3D Resnet18,DFEW fd2            &\textbf{53.00}\\
3D Resnet18,EC-STFL,DFEW fd2    &\textbf{53.26}\\
\bottomrule
\end{tabular}
\end{center}
\end{table}

\section{Conclusions and Future work}
In this paper, we have presented a new large-scale unconstrained dynamic facial expression database, DFEW, and proposed a novel spatiotemporal deep feature learning framework, EC-STFL, to deal with dynamic FER in the wild. To the best of our knowledge, our DFEW has the largest number of samples compared with existing databases of dynamic facial expression in the wild, which containing 16,372 video clips extracted from over 1500 different movies. More importantly, DFEW has provided the reliable distribution information of 7 basic expressions for all the video clips because 10 well-trained annotators independently annotate each sample of DFEW. We also conducted extensive baseline experiments on DFEW under the well-designed protocol by using well-performing spatiotemporal deep learning methods as well as the proposed EC-STFL framework and deeply discussed the results. Experimental results showed that our DFEW is a promising unconstrained dynamic facial expression database and the proposed EC-STFL framework can improve the performance of spatiotemporal deep neural networks in coping with dynamic FER in the wild. In the future, we will continue to maintain DFEW by collecting more samples and providing more types of label information such that DFEW can better promote the progress of FER research.

\begin{acks}
This work was supported in part by the National Key Research and Development Program of China under Grant 2018YFB1305200, in part by the National Natural Science Foundation of China under Grant 61921004, Grant 61902064, and Grant 81971282, and in part by the Fundamental Research Funds for the Central Universities under Grant 2242018K3DN01.  
\end{acks}

\bibliographystyle{ACM-Reference-Format}
\bibliography{MM2020_final}

\end{document}